\documentclass[conference,letterpaper]{IEEEtran}

\usepackage[utf8]{inputenc} 
\usepackage[T1]{fontenc}
\usepackage{url}
\usepackage{ifthen}
\usepackage{cite}
\usepackage[cmex10]{amsmath} 

\usepackage{xcolor}

\usepackage{graphicx} 
\usepackage{amsfonts} 
\usepackage{amssymb} 
\newtheorem{theorem}{\textbf{Theorem}}[section]
\newtheorem{corollary}{\textbf{Corollary}}[theorem]
\newtheorem{lemma}[theorem]{\textbf{Lemma}}
\newtheorem{definition}{\textbf{Definition}}[section]

\newcommand{\eq}[1]{\begin{align*}#1\end{align*}}

\newcommand{\beq}{\begin{equation}}
\newcommand{\eeq}{\end{equation}}

\begin{document}
\title{Shallow Neural Network can Perfectly Classify an Object following Separable Probability Distribution} 

\author{%
  \IEEEauthorblockN{Youngjae Min and Hye Won Chung}
  \IEEEauthorblockA{School of Electrical Engineering\\
                    KAIST\\
                    Email: \{yjmin313, hwchung\}@kaist.ac.kr}
}

\maketitle

\begin{abstract}
Guiding the design of neural networks is of great importance to save enormous resources consumed on empirical decisions of architectural parameters. This paper constructs shallow sigmoid-type neural networks that achieve 100\% accuracy in classification for datasets following a linear separability condition. The separability condition in this work is more relaxed than the widely used linear separability. 
Moreover, the constructed neural network guarantees perfect classification for any datasets sampled from a separable probability distribution.
This generalization capability comes from the saturation of sigmoid function that exploits small margins near the boundaries of intervals formed by the separable probability distribution. 
\footnote{This work was supported by National Research Foundation of Korea under grant number 2017R1E1A1A01076340 and by the Ministry of Science and ICT, Korea, under the ITRC support program (IITP-2019-2018-0-01402). }
\end{abstract}


\section{Introduction}

Learning with neural networks has shown remarkable performance in a variety of tasks. 
To explain the great success of neural networks, there have been many attempts to analyze the performance in terms of architectural parameters, training dataset and algorithms. However, there are still many open problems, especially in guiding the design of neural networks.
Our focus is on developing guidance for construction of a neural network for classification tasks with proper architectural parameters that generalizes well to any datasets sampled from a separable probability distribution. 


Selecting architectural parameters of a neural network has been a challenging problem and enormous resources have been consumed to empirically choose the best structure. 
To guide the design of neural networks, there have been efforts to understand the expressivity of neural networks; 
the works in~\cite{func_hornik,func_leshno,func_ferrari} have investigated the functional capability of neural networks and the works in~\cite{lr_bengio,lr_montufar,lr_serra} have studied the characteristics of linear regions formed by the decision boundary of a neural network. While these approaches analyze given architectures without consideration of datasets, we focus on finding proper architectural parameters of a neural network based on characteristics of datasets. This work is inspired by \cite{finitework} where an upper bound on the number of parameters is investigated to perfectly fit finite training dataset satisfying a particular separability condition. Despite of the perfect fitting, the constructed networks does not guarantee any generalization beyond the training datasets, even though the authors appeal to the philosophy of Occam's razor to argue that the networks' parsimonious structure is likely to generalize well. In our work, we fill this gap and design a network that generalizes to any datasets sampled from a separable probability distribution.

The main idea is to use sigmoid-type neural networks (instead of ReLU-type used in~\cite{finitework}) in which the saturation of the activation function can be utilized to design networks that generalize well. 
In particular, we first deal with linearly separable data of which similar concepts are introduced in previous works, \cite{finitework,sgd_brut,sep_soudry}. 
For this linearly separable data, we can find a projection vector such that projecting the data onto the vector creates separable intervals each containing data of the same class.
For such a dataset, by scaling the small margins between the intervals, we design a network whose activation function is saturated and guarantees the similar patterns for any data points in the same interval. This design technique leads to generalization of the neural network (Sec.~\ref{sec:1D}).
Then, we extend the concept of separability through multiple projections and construct 4-layer neural networks that perfectly classify data satisfying the extended separability condition (Sec.~\ref{sec:multi}).


This work introduces an interesting aspect for generalization of neural networks. While recent studies, \cite{sep_soudry,ovp_zhang,ovp_li,ovp_du}, on generalization focus on ReLU-type neural networks and over-parameterization, we employ sigmoid function and show that its saturation effect guarantees generalization to the separable distribution of data even with parsimonious and shallow network structure. 
Avoiding the vanishing gradient problem~\cite{vanishing} of deep neural networks, this work demonstrates the potential benefits of shallow sigmoid-type neural networks in generalization.

\subsection{Prior Work}

In~\cite{finitework}, the authors investigate how to construct a neural network to exactly fit the finite training data $\{(x_i,y_i)\}_{i=1}^p$ with $x_i\in\mathbb{R}^d$ and $y_i\in[1:c]$ where the data satisfies `$s$-separability.' The separability means that there exists a vector $a\in\mathbb{R}^d$ such that projecting the data $\{x_i\}_{i=1}^p$ onto $a$ creates $(s+1)$ separate intervals each containing points with the same label $y\in[1:c]$.
Assuming the intervals have $k_1, k_2, \dots, k_{s+1}$ data points, respectively, and the dataset is ordered properly, the separability condition implies that there exist boundaries $\{b(1),\dots,b(s+1)\}$ such that
\beq\label{eqn:sep_prior}
\begin{split}
&b(1)<a^Tx_1<a^Tx_2<\cdots<a^Tx_{k_1}<b(2)\\
&<a^Tx_{k_1+1}<\cdots<a^Tx_{k_1+k_2}<b(3)\\
&<\cdots<b(s+1)<a^Tx_{p-k_{s+1}+1}<\cdots<a^Tx_p.
\end{split}
\eeq

When $f$ is an injective mapping from the class labels to $\mathbb{R}^m$, a 2-layer neural network  is constructed with $(d+(m+1)s)$ parameters for $(s-1)$-separable dataset that maps the training data with label $j$ to $f(j)$.
The hidden layer is composed of ReLU activation function which outputs the max of its input and 0.

The work of~\cite{finitework} answers to the question of how many free parameters are sufficient to fit the training data exactly when the data satisfies the particular separability condition in~\eqref{eqn:sep_prior}. However, it does not answer whether or not such a network can generalize well to test data. 
In our work, we fill this gap.

\section{Network Construction for Classification of Separable Dataset with Margins}\label{sec:1D}

In this section, we generalize the concept of $k$-separability by introducing $\delta$-margin between intervals and find an exact construction of a neural network that can not only fit a training dataset exactly but can also generalize well to any test datasets sampled from a separable probability distribution. 

\begin{figure}[!t]
\centering
\includegraphics[width=2.5in]{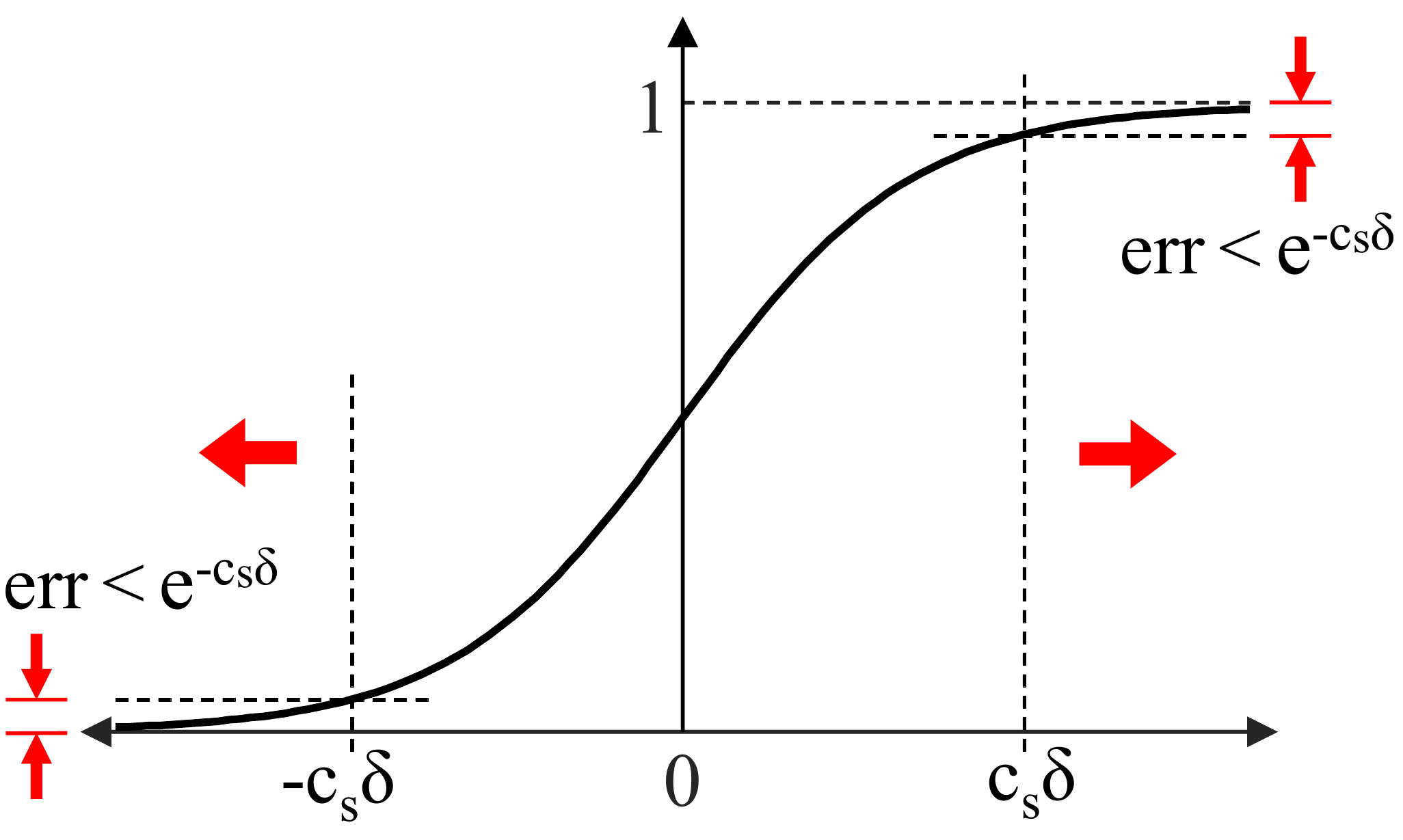}
\caption{Saturation of sigmoid function through scaling of $\delta$-margin.}
\label{fig_sigmoid}
\end{figure}

The main idea is to use sigmoid-type neural network instead of ReLU-type to utilize saturation effect by scaling weights of neural network. 
Unlike ReLU function used in \cite{finitework}, sigmoid function, defined as
\beq\label{eqn:sigmoid}
\rho(t)={1}/({1+e^{-t}}),
\eeq
converges to $1$ as $t\to\infty$ and to 0 as $t\to-\infty$ as shown in Fig.~\ref{fig_sigmoid}. This saturation effect is exploited to construct a neural network that generalizes to test dataset.  

To elaborate the idea, we first introduce the definition of $k$-separability with $\delta$-margin.


\begin{definition}\label{def:sep}
Let $\mathcal{X} \subset \mathbb{R}^{d}$ and $\mathcal{Y}=[1:c]$. A distribution \(D\) over \(\mathcal{X} \times \mathcal{Y}\) is \(k\)-separable with \(\delta\)-margin (for some $\delta>0$) if there exist a projection vector \(a \in \mathbb{R}^{d}\) with $\|a\|_2=1$ and constants \(b_1<b_2<\cdots<b_{k+1}\) such that, for $\mathcal{X}_i:=\{x\in \mathcal{X}:b_i+\delta<a^Tx<b_{i+1}-\delta\}$, $i\in[1:k]$,
\renewcommand{\labelenumi}{(\roman{enumi})}
\begin{enumerate}
\item \(\mathbb{P}_{(x,y)\sim D}\left(y=y_i \mid \mathcal{X}_i\right) = 1\) for some \(y_i \in \mathcal{Y}\),
\item \(\mathbb{P}_{(x,y)\sim D}\left(\bigcup_{i=1}^{k}\mathcal{X}_i\right)= 1\).
\end{enumerate}
\end{definition}
Note that the number $k$ of intervals can be larger than the number $c$ of labels, which means that all data of the same label needs not be in only one interval. 


For any $x\in \mathcal{X}_i$, $|a^T x-b_l|$ is bounded below by $\delta$ for all $l\in[1:k]$. 
When we scale the weight $a$ and bias $b$ by a large constant $c_s$, the output of the sigmoid function $\rho\left(c_s(a^Tx-b_l)\right)$ converges to 1 for $l\leq i$ and to 0 for $l>i$ as $c_s\to\infty$. 
Therefore, this saturation effect of sigmoid function can guarantee the same output value for any data $x$ in the same interval $\mathcal{X}_i$.
We utilize this effect to construct a 2-layer neural network that can classify any datasets sampled from a $k$-separable distribution with $\delta$-margin as defined in Def.~\ref{def:sep}.
We also give an upper bound on the scaling factor $c_s$ that allows a small fixed error $\epsilon>0$ in the desired output of the neural network. 



\begin{figure}[!t]
\centering
\includegraphics[width=2.5in]{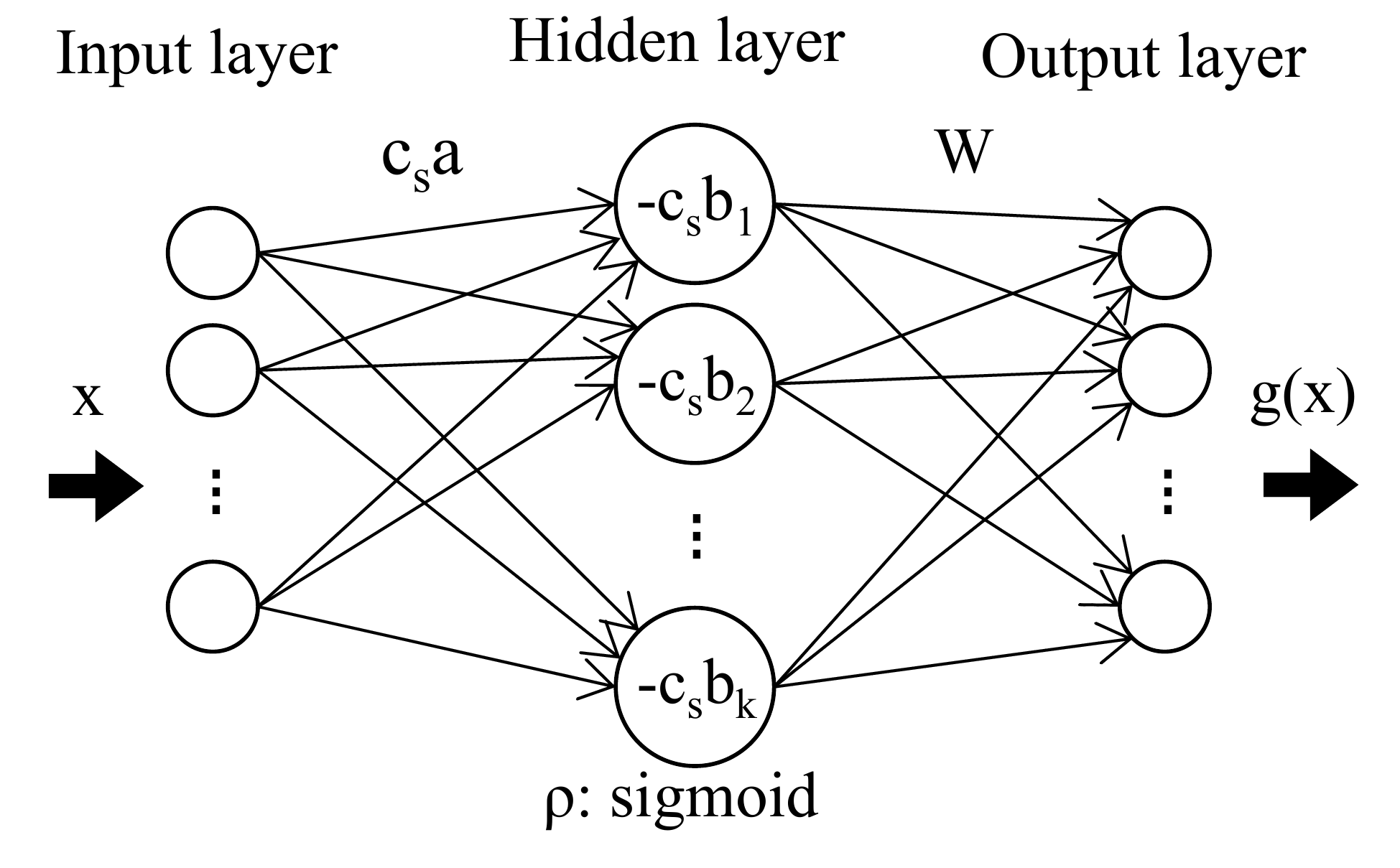}
\caption{Constructed neural network that outputs $g(x)$ arbitrarily close to the desired output of input $x$ when the input is sampled from a $k$-separable distribution with $\delta$-margin.}
\label{fig_2layer}
\end{figure}

\begin{theorem}
\label{thm:main1}
{\it
Let a distribution \(D\) be \(k\)-separable with \(\delta\)-margin by a projection vector $a\in\mathbb{R}^d$ as defined in  Def.~\ref{def:sep} and \(f\) be an injective function that maps the label set \(\mathcal{Y}\) into \(\mathbb{R}^{m}\). For \((x,y)\in \mathcal{X}\times \mathcal{Y}\), consider \(f(y)\in\mathbb{R}^m\) as the desired output of a neural network for input $x$ with label $y$. Then, for any \(\epsilon>0\), the 2-layer neural network, \(g:\mathcal{X}\rightarrow\mathbb{R}^{m}\), shown in Fig.~\ref{fig_2layer} with parameters $a \in\mathbb{R}^d$ (the projection vector), $\{b_1,\dots, b_k\}$ (the boundary of intervals), 
\begin{align}
& {W} = 
\begin{bmatrix}
f(y_1)^T\\
f(y_2)^T-f(y_1)^T\\
\vdots\\
f(y_k)^T-f(y_{k-1})^T
\end{bmatrix}
=[w_1\;w_2\;\cdots\;w_m],
\label{eqn:W}\\
&
c_s=(1/\delta)\log\left(\left(\sqrt{k}\cdot \left(\max_{1\leq j\leq m}\|w_j\|_2\right)\right)/{\epsilon}\right)
\label{eqn:cs}
\end{align}
satisfies
\beq\label{eqn:accuracy}
\mathbb{P}_{(x,y)\sim D}\left(\max_{1\leq j\leq m}|g_j(x)-f_j(y)|>\epsilon\right)=0
\eeq
where \(f_j\) and \(g_j\) denote the $j$-th components of \(f\) and \(g\), respectively.
This network is specified by total \((d+(m+1)k)\) parameters.
}
\end{theorem}
\IEEEproof{
We prove that for any $(x,y)\sim D$ the constructed 2-layer neural network in Fig.~\ref{fig_2layer} with parameters $(c_s,a,\{b_1,\dots, b_k\}, {W})$ outputs $g(x)$ that is close to the desired output $f(y)$ in the sense of~\eqref{eqn:accuracy}.

Let the vector $h_i\in \{0,1\}^k$ represent the saturated output of the hidden layer when the sigmoid function $\rho$ in~\eqref{eqn:sigmoid} is applied to $c_s(a^Tx- b_l)$, $1\leq l\leq k$, for any input $x\in \mathcal{X}_i$ and  the scaling factor $c_s\to\infty$.
Note that $h_i=[\underbrace{1\dots1}_{i} \underbrace{0 \dots 0}_{k-i}]$. Let $H\in \{0,1\}^{k\times k}$ be the matrix with its $i$-th row equal to $h_i^T$, i.e, 
\beq
{H}=
\begin{bmatrix}
1 & 0 & \cdots & 0\\
1 & 1 & \cdots & 0\\
\vdots & \vdots & \ddots & \vdots\\
1 & 1 & \cdots & 1
\end{bmatrix}
= \begin{bmatrix}
-h_1^T-\\
-h_2^T-\\
\vdots\\
-h_k^T-
\end{bmatrix}.
\eeq
We next construct the weight matrix $W\in \mathbb{R}^{k\times m}$ of the output layer such that $HW=Y$ where $Y\in \mathbb{R}^{k\times m}$ is the matrix whose $i$-th row is the desired output $f(y_i)$ for the label $y_i\in[1:c]$ for input $x\in \mathcal{X}_i$. 
Since $H$ is invertible, $W$ that satisfies $HW=Y$ is $W=H^{-1}Y$ which is equal to~\eqref{eqn:W}.
Therefore, when $c_s\to\infty$, the constructed 2-layer neural network outputs the desired output $f(y_i)$ for any input $x\in \mathcal{X}_i$ for $i\in[1:k]$.

We next analyze an upper bound on the scaling factor $c_s$ when we allow size-$\epsilon$ error between the actual output $g_j(x)$ and the desired output $f_j(y)$ for $1\leq j\leq m$ when $(x,y)\sim D$. 

Note that the output of the sigmoid function is bounded below and above by
\beq
1-e^{-t}<\rho(t)=1/({1+e^{-t}})<e^{t}.
\eeq
By using this and the definition of the $k$-separable distribution with $\delta$-margin, we can bound the input and output of the $l$-th neuron of the hidden layer as below.
For $x\in \mathcal{X}_i$,
\renewcommand{\labelenumi}{(\roman{enumi})}
\begin{enumerate}
\item if \(l\leq i\),
\vspace{-5pt}
\beq\label{eqn:bd1layer1}
\begin{split}
& c_sa^Tx - c_sb_l \geq c_sa^Tx - c_sb_i > c_s\delta,\\
& \rho(c_sa^Tx{-}c_sb_l){>}1{-}e^{-(c_sa^Tx - c_sb_l)}{>}1{-}e^{-c_s\delta}.
\end{split}
\eeq
\item if \(l>i\)
\vspace{-5pt}
\beq\label{eqn:bd1layer2}
\begin{split}
& c_sa^Tx - c_sb_l \leq c_sa^Tx - c_sb_{i+1} < -c_s\delta,\\
& \rho(c_sa^Tx - c_sb_l)<e^{(c_sa^Tx - c_sb_l)}<e^{-c_s\delta}.
\end{split}
\eeq
\end{enumerate}
When we denote the output of the hidden layer for an input $x\in\mathcal{X}_i$ by $(h_i+n)\in \mathbb{R}^k$,
the output of the output layer is
\beq
g(x)={W}^T({h_i}+n)=f(y_i)+{W}^Tn.
\eeq
Moreover, as shown by~\eqref{eqn:bd1layer1} and~\eqref{eqn:bd1layer2}, $|n_l|<e^{-c_s\delta}$ for all $l\in[1:k]$.
Thus, we can bound
\beq
|g_j(x)-f_j(y_i)|=|w_j^T n|\leq\|w_j\|_2\cdot \|n\|_2\leq \|w_j\|_2\sqrt{k} e^{-c_s\delta}
\eeq
where the first inequality is by Cauchy–Schwarz inequality.
To guarantee $|g_j(x)-f_j(y_i)|\leq \epsilon$ for a fixed $\epsilon>0$ for all $1\leq j\leq m$, it is sufficient to have the scaling factor $c_s$ in~\eqref{eqn:cs}.

Expressing \(\mathbb{P}_{(x,y)\sim D}\) simply as \(\mathbb{P}\),
\beq
\begin{split}
&\mathbb{P}\left(\max_{1\leq j\leq m}|g_j(x)-f_j(y)|>\epsilon\right)\\
&=\sum_{i=1}^{k}\mathbb{P}\left(\max_j|g_j(x){-}f_j(y_i)|>\epsilon \Big| \mathcal{X}_i\right)\mathbb{P}(\mathcal{X}_i)
= 0 
\end{split}
\eeq
where the first equality is from Def.~\ref{def:sep} and the second equality is from  $|g_j(x)-f_j(y_i)|\leq\epsilon, \;\forall x\in \mathcal{X}_i,\;\forall j$ with the chosen $c_s$.

The constructed neural network requires $d$ parameters for $c_sa\in \mathbb{R}^d$, \(k\) for the hidden layer's biases $(c_sb_1,\dots,c_sb_k)$, and \(mk\)  for the output layer's weight matrix $W\in\mathbb{R}^{k\times m}$.\;\IEEEQED
}

As an example, when one-hot encoding is used for classification, there are $c$ nodes in the output layer and the estimate of the label for an input to the network is chosen by finding the output node of the largest value.
Then, for $f$ being the one-hot encoding of $\mathcal{Y}$, the network constructed in Thm.~\ref{thm:main1} guarantees the perfect classification of $(x,y)\sim D$ when the scaling factor $c_s$ in~\eqref{eqn:cs} is chosen for $\epsilon=1/2$. 


\begin{corollary}\label{cor:1}
{\it
For any data sampled from a distribution $D$ that is \(k\)-separable with \(\delta\)-margin as defined in Def.~\ref{def:sep}, perfect classification is possible with a 2-layer neural network in Fig.~\ref{fig_2layer}  with \((d+(c+1)k)\) parameters. 
}
\end{corollary}

\subsection{Simulation Results}

\begin{figure}[!t]
\centering
\includegraphics[width=8cm]{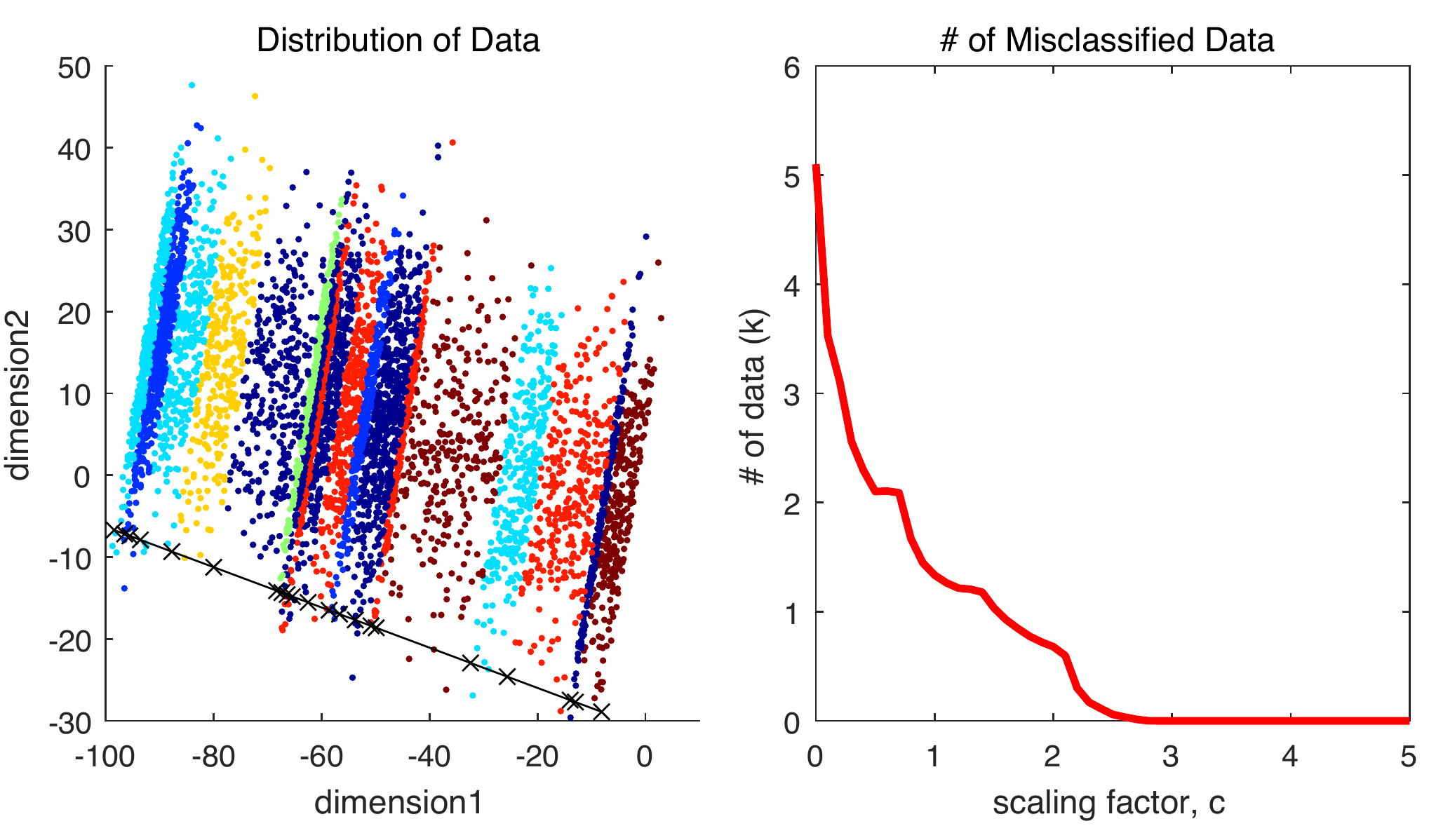}
\caption{6k samples from a 20-separable distribution with 0.1-margin (left). With a neural network constructed as in Thm.~\ref{thm:main1}, the number of misclassified data is counted varying the scaling factor, $c_s$ (right).}
\label{fig_scaling}
\end{figure}

We present a simulation result that shows the effect of saturation of sigmoid function on the classification performance.
In Fig~\ref{fig_scaling}, for $\mathcal{X}\subset\mathbb{R}^{2}$ and $\mathcal{Y}=[1:10]$,
a dataset following a 20-separable distribution with 0.1-margin is generated with random \(a\) and \(\{b_i\}\).
Then, we construct a  2-layer neural network in Fig.~\ref{fig_2layer} with a different scaling factor $c_s>0$ to observe the effect of saturation on the classification performance.
Simulation result shows that the number of misclassified data decreases as $c_s$ increases and drops to 0 for $c_s\geq 3$. 
This value is smaller than the sufficient $c_s$, $11.02$, calculated by~\eqref{eqn:cs} in Thm.~\ref{thm:main1} for $\epsilon=1/2$.



\section{Extension to Multi-Dimensional Projection}\label{sec:multi}

We extend the result in Sec.~\ref{sec:1D} by generalizing the separability and considering datasets that can be separable by $n$ projection vectors instead of by one. For the broader class of datasets, we design a 4-layer neural network that can perfectly classify any datasets following the generalized separability. 


\begin{definition}\label{def:sep2}
Let $\mathcal{X}\subset\mathbb{R}^{d}$ and $\mathcal{Y}{=}[1:c]$. A distribution \(D\) over \(\mathcal{X} \times \mathcal{Y}\) is \((k_1,k_2,\cdots,k_n)\)-separable with \(\delta\)-margin (for some $\delta>0$) if there exist projection vectors \(a_1,a_2,\cdots,a_n \in \mathbb{R}^{d}\) with $\|a_s\|_2=1$ and constants $b_{s,1}{<}b_{s,2}{<}\cdots{<}b_{s,k_s+1}$ for $1{\leq}s{\leq}n$, such that, for \(\mathcal{X}_{\mathbf{i}}{=}\{x\in \mathcal{X}:b_{s,i_s}{+}\delta<a_s^Tx<b_{s,i_s+1}{-}\delta\;\text{for}\;1{\leq}s{\leq}n\}\), \(\mathbf{i}{=}(i_1,i_2,\cdots,i_n)\), with \(i_s\in[1:k_s]\;\text{for}\;1{\leq}s{\leq}n\),
\renewcommand{\labelenumi}{(\roman{enumi})}
\begin{enumerate}
\item \(\mathbb{P}_{(x,y)\sim D}\left(y=y_\mathbf{i} \mid \mathcal{X}_\mathbf{i}\right) = 1\) for some \(y_\mathbf{i} \in \mathcal{Y}\),
\item \(\mathbb{P}_{(x,y)\sim D}\left(\bigcup_\mathbf{i}\mathcal{X}_\mathbf{i}\right)= 1\).
\end{enumerate}
\end{definition}

\begin{figure}[!t]
\centering
\includegraphics[width=6cm]{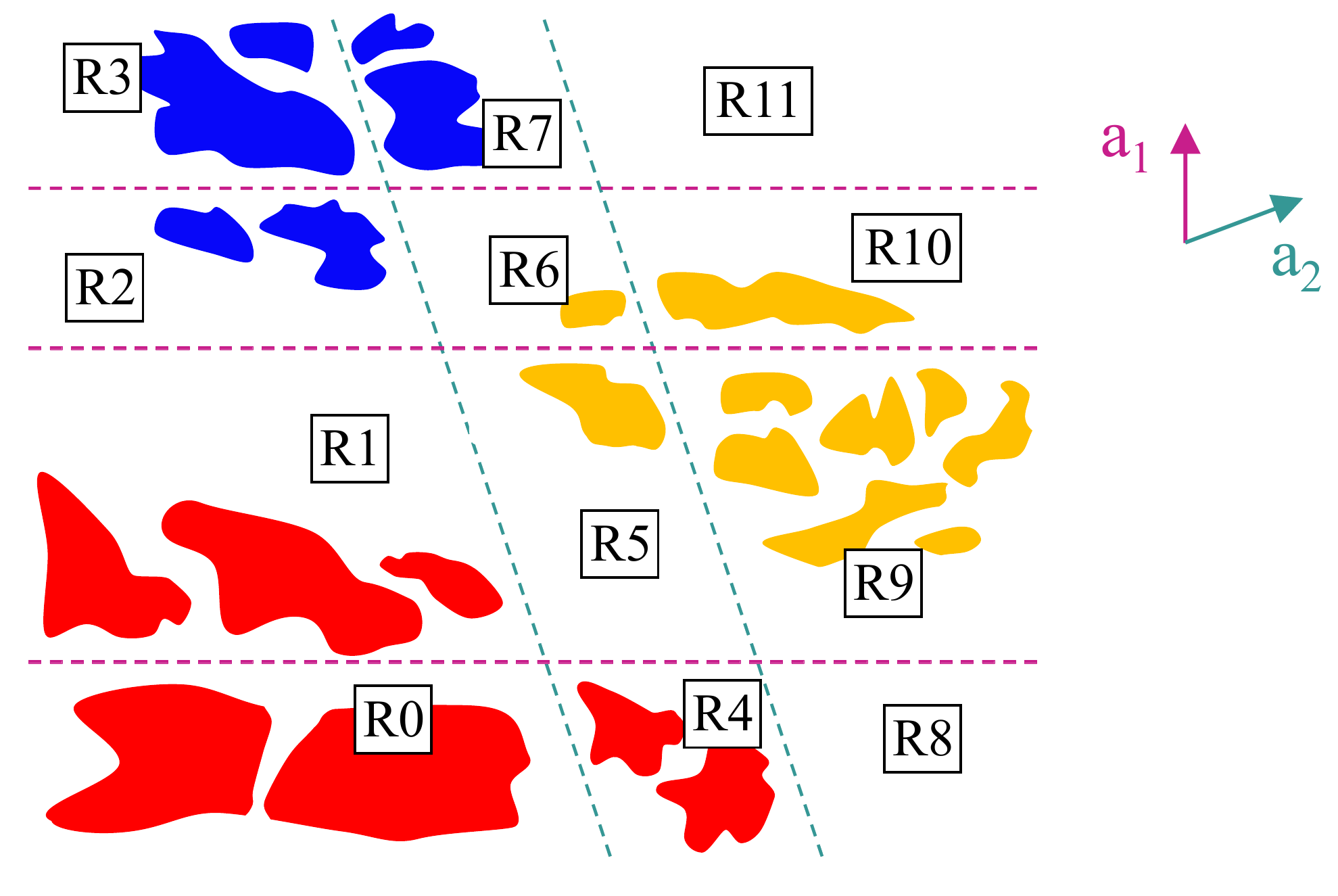}
\caption{Distribution of data which is separable by two projection vectors. The area with nonzero probability density is expressed using different colors for each class.}
\label{fig_2d-sep}
\end{figure}

As an example, consider the 2D dataset in Fig.~\ref{fig_2d-sep}. This dataset does not have a vector to which its projection is separable, but the two vectors, \(a_1\) and \(a_2\), can separate the data into 12 different regions. 

\begin{figure}[!t]
\centering
\includegraphics[width=2.5in]{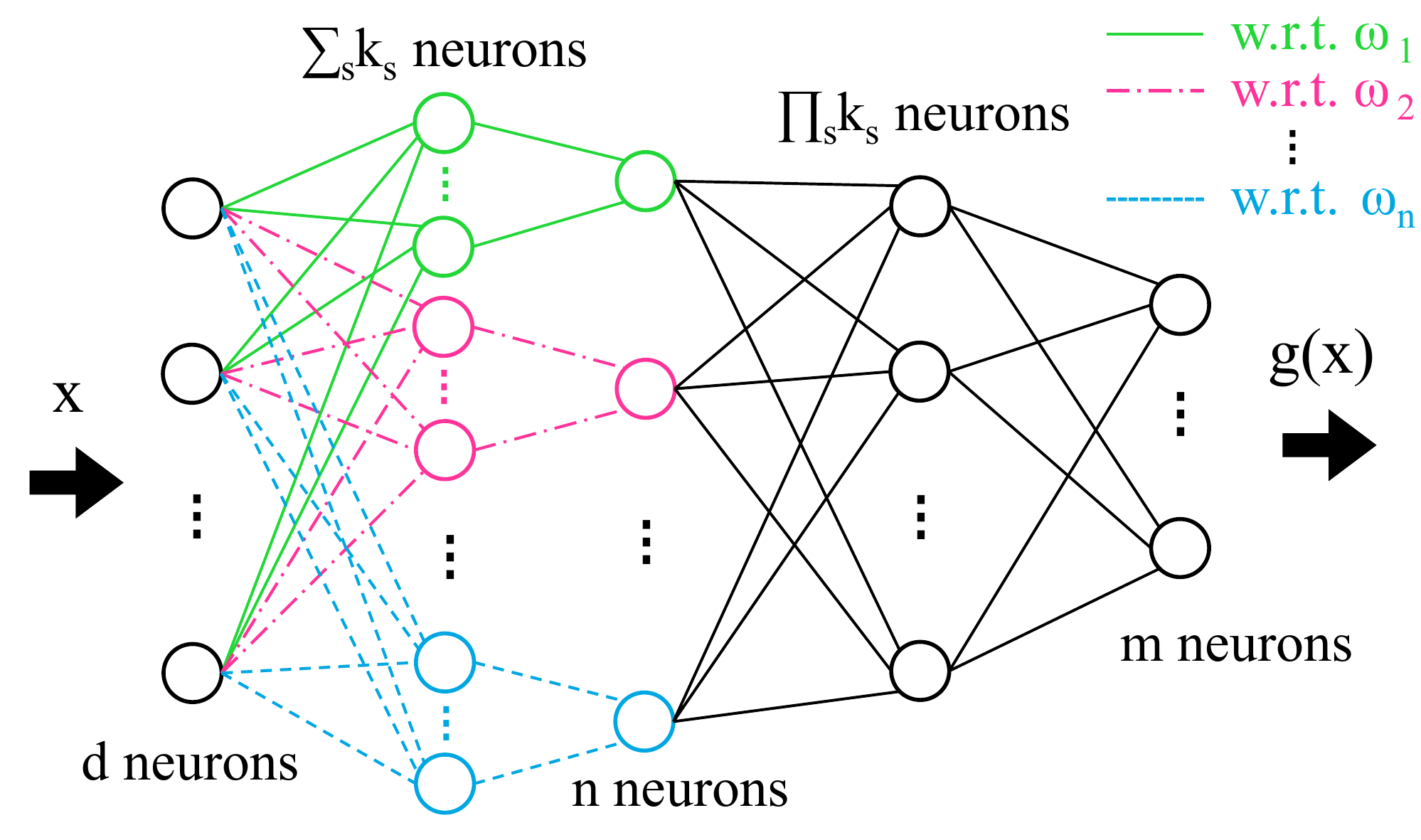}
\caption{Constructed 4-layer neural network that outputs $g(x)$ arbitrarily close to the desired output of input $x$ sampled from a $(k_1,\dots,k_n)$-separable distribution with $\delta$-margin, separable by $n$ projection vectors $\{a_1,\dots,a_n\}$.}
\label{fig_4layer}
\end{figure}

We construct a 4-layer neural network shown in Fig.~\ref{fig_4layer} that can classify this type of broader class of separable dataset.
The first two hidden layers consist of $n$ subnetworks (differentiated by its color) each of which is constructed by the 2-layer neural network (with only one output node) as in Thm.~\ref{thm:main1} with respect to each projection vector $a_s\in\mathbb{R}^d$, $1\leq s\leq n$. 
The role of the first two layers is to map the original dataset into a new dataset satisfying the 1D separability in Def.~\ref{def:sep}, i.e., the dataset that is $\left(\prod_{s=1}^n k_s\right)$-separable by a projection vector $a\in\mathbb{R}^n$ with $\frac{1}{4\sqrt{n}}$-margin. 
Once the dataset is transformed to a separable dataset satisfying Def.~\ref{def:sep}, by Thm.~\ref{thm:main1} it can be fed into a 2-layer neural network (the last two layers in Fig.~\ref{fig_4layer}) to have desired outputs. 
The role of the first two layers in the neural network is stated in the following lemma.


\begin{lemma}
{\it
For data $(x,y)$ following a distribution $D$ that is $(k_1,k_2,\dots,k_n)$-separable with $\delta$-margin by $n$ projection vectors $(a_1,\dots,a_n)$ as defined in Def~\ref{def:sep2},
there exists a 2-layer network (the first 2 layers in Fig.~\ref{fig_4layer}) that implements $p: \mathcal{X}\to \mathbb{R}^n$
such that $(p(x),y)$ follows a distribution $D'$ that is $\left(\prod_{s=1}^n k_s\right)$-separable with $\left(\frac{1}{4\sqrt{n}}\right)$-margin by a projection vector $a=\frac{1}{\sqrt{n}}[1,1,\dots,1]^T\in\mathbb{R}^n$. 
}\label{lem:map_sep}
\end{lemma}
\IEEEproof{
The full proof is stated in Appendix. 
Here we only provide a proof sketch.

For \(s\in[1:n]\), let 
\beq\label{eqn:hs}
h^s(i):=(i-1)\prod_{p=0}^{s-1}k_p, \text{ for }i\in[1:k_s], \text{ with }k_0=1.
\eeq 
For each $s\in[1:n]$, we construct a 2-layer subnetwork, $p^s:\mathcal{X}\to \mathbb{R}$, whose output $p^s(x)$ for input $x\in  \mathcal{X}_{\mathbf{i}}$, $\mathbf{i}=(i_1,\dots,i_n)$, is close to the desired output $h^s(i_s)$. Such a subnetwork can be constructed by using Thm.~\ref{thm:main1} for each projection vector $a_s\in\mathbb{R}^d$.  
All such $n$ subnetworks are combined in parallel to form the first two layers of the network shown in~Fig.~\ref{fig_4layer}, whose output is $p(x)=[p^1(x),  p^2(x),  \cdots  ,p^n(x)]^T\in\mathbb{R}^n$.

Note that by Thm.~\ref{thm:main1}, with the properly chosen scaling constant $c_s$ in~\eqref{eqn:cs} we can make each output $p^s(x)$ close to its desired value $h^s(i_s)$. 
Moreover, due to the definition of $h^s(i)$ in~\eqref{eqn:hs}, $\sum_{s=1}^n h^s (i_s)$ for each $\mathbf{i}=(i_1,\dots, i_n)$ has different integer value.
Therefore, for a projection vector $a=\frac{1}{\sqrt{n}}[1, 1, \cdots , 1]^T\in\mathbb{R}^n$, we can make $a^T p(x)=\frac{1}{\sqrt{n}}\sum_{s=1}^n p^s (i_s)$ for $x\in\mathcal{X}_{\mathbf{i}}$ be different  for each $\mathbf{i}=(i_1,\dots, i_n)$. Since there are $\left(\prod_{s=1}^n k_s\right)$ different $\mathbf{i}$'s, this implies that $(p(x),y)$ follows a distribution $D'$ that is $\left(\prod_{s=1}^n k_s\right)$-separable by $a=\frac{1}{\sqrt{n}}[1, 1, \cdots , 1]^T$ with some margin. 
\IEEEQED



}

After mapping the original dataset $(x,y)$ to a new dataset $(p(x),y)$ that is separable by one projection vector $a\in\mathbb{R}^n$ as in Lemma~\ref{lem:map_sep}, we can construct the next 2-layer network (the last two layers in Fig.~\ref{fig_4layer}) to map the new dataset to any desired output for classification by using Thm.~\ref{thm:main1}.

\begin{theorem}
{\it
Let \(D\) be a \((k_1,k_2,\cdots,k_n)\)-separable with \(\delta\)-margin as defined in  Def.~\ref{def:sep2}  and \(f\) be an injective function that maps \(\mathcal{Y}\) into \(\mathbb{R}^{m}\). For \((x,y)\in \mathcal{X}\times \mathcal{Y}\), consider \(f(y)\) as the desired output of a neural network for input $x$ with label $y$. Then, for any \(\epsilon>0\), there exists a 4-layer neural network, \(g:\mathcal{X}\rightarrow\mathbb{R}^{m}\), shown in Fig.~\ref{fig_4layer}, with \((n(d+1)+2\sum_{s=1}^{n}k_s+(m+1)\prod_{s=1}^{n}k_s)\) parameters such that
\beq
\mathbb{P}_{(x,y)\sim D}\left(\max_{1\leq j\leq m}|g_j(x)-f_j(y)|>\epsilon\right)=0
\eeq
where \(f_j\) and \(g_j\) denote the $j$-th components of \(f\) and \(g\), respectively.
}\label{thm:main2}
\end{theorem}

Here, note that \(\sum_{s=1}^{n}(d+2k_s)\) parameters are required for the first two layers, and \((n+(m+1)\prod_{s=1}^{n}k_s)\) for the last two layers by Thm~\ref{thm:main1}.

By the similar reasoning as in Cor~\ref{cor:1}, Thm.~\ref{thm:main2} implies the perfect classification 
for $D$ satisfying Def.~\ref{def:sep2}.

\begin{corollary}\label{cor:2}
{\it
For data following a distribution $D$ that is \((k_1,k_2,\cdots,k_n)\)-separable with \(\delta\)-margin as defined in Def.~\ref{def:sep2}, perfect classification is possible with a 4-layer neural network in Fig.~\ref{fig_4layer} with  \((n(d+1)+2\sum_{s=1}^{n}k_s+(c+1)\prod_{s=1}^{n}k_s)\) parameters.
}
\end{corollary}

\section{Conclusion}

We construct shallow sigmoid-type neural networks that achieve perfect classification for separable dataset with margins. This shows the benefit of saturation of sigmoid function in generalization, in the sense that it can generalize to the distribution of data beyond finite samples under a separability condition. The results suggest further investigation on the generalization of shallow sigmoid-type neural networks. There are also remaining open problems such as finding proper projection vectors and boundaries for a given separable dataset and approximating a general dataset to a separable one. 


\appendix

\section*{Proof of Lemma~\ref{lem:map_sep}}

For \(s\in[1:n]\), let \(\mathcal{Y}_s=[1:k_s]\) and \(D_s\) be a replication of \(D\), except changing \(\mathcal{Y}\) to \(\mathcal{Y}_s\) and letting \(y_\mathbf{i}=i_s\) for $\mathbf{i}=(i_1,i_2,\dots,i_n)$.
Since $D_s$ is $k_s$-separable with $\delta$-margin, by Thm.~\ref{thm:main1} a 2-layer network, $p^s:\mathcal{X}\to R$, can be constructed with $(d+2k_s)$ parameters and make
$
\mathbb{P}_{(x,y)\sim D_s}\left(|p^s(x)-h^s(y)|>\frac{1}{4n}\right)=0.
$
This also implies that 
\beq\label{eqn:clue1}
\mathbb{P}_{(x,y)\sim D_s}\left(|p^s(x)-h^s(i_s)|\leq\frac{1}{4n}\Big|\mathcal{X}_\mathbf{i}\right)=1
\eeq
since for $x\in \mathcal{X}_\mathbf{i}$ the label $y=i_s$ under the distribution $D_s$.

For $p(x)=[p^1(x),  p^2(x),  \cdots  ,p^n(x)]^T\in\mathbb{R}^n$, let $D'$ be the distribution of $(p(x),y)$ where $(x,y)\sim D$. 
We will show that $D'$ is $(\prod_{s=1}^n k_s)$-separable with $\left(\frac{1}{4\sqrt{n}}\right)$-margin by a projection vector $a=\frac{1}{\sqrt{n}}[1,1,\dots,1]^T\in\mathbb{R}^n$.

For \(\mathbf{i}=(i_1,i_2,\cdots,i_n)\), define \(\tilde{k}{:}\prod_{s=1}^{n}[1:k_s]\rightarrow[0:\prod_{s=1}^{n}k_s-1]\) as 
\beq
\tilde{k}(\mathbf{i})=\sum_{s=1}^{n}h^s(i_s)
\eeq
 for $h^s(i)$ in~\eqref{eqn:hs}. 
Note that $\tilde{k}$ is bijective (one-to-one and onto).
Expressing \(\mathbb{P}_{(p(x),y)\sim D'}\) simply as \(\mathbb{P}\),
\beq
\begin{split}\nonumber
1=&\mathbb{P}\left(|p^1(x){-}h^1(i_1)|{\leq}\frac{1}{4n},\cdots,|p^n(x){-}h^n(i_n)|{\leq}\frac{1}{4n} \Big| \mathcal{X}_{\mathbf{i}}\right)\\
\leq & \; \mathbb{P}\left(\left|\sum_{s}(p^s(x)-h^s(i_s))\right|\leq\frac{1}{4} \Big| \mathcal{X}_{\mathbf{i}}\right)\\
=&\; \mathbb{P}\left(\left|a^Tp(x)-\frac{\tilde{k}(\mathbf{i})}{\sqrt{n}}\right|\leq\frac{1}{4\sqrt{n}} \Big| \mathcal{X}_{\mathbf{i}}\right)=1.
\end{split}
\eeq
Therefore, by defining the boundaries $b_i=\frac{i-1.5}{\sqrt{n}}$ for $i\in[1:\prod_{s}k_s+1]$,
\beq\label{eqn:rel1}
\mathbb{P}\left(b_{\tilde{k}(\mathbf{i})+1}{+}\frac{1}{4\sqrt{n}}<a^Tp(x)<b_{\tilde{k}(\mathbf{i})+2}{-}\frac{1}{4\sqrt{n}} \Big| \mathcal{X}_{\mathbf{i}}\right)=1.
\eeq

By using~\eqref{eqn:rel1} and the fact that the original distribution $D$ of $(x,y)$ satisfies Def.~\ref{def:sep2}, we show that for $\mathcal{X}^k:=\{x\in \mathcal{X}:b_{k+1}+\frac{1}{4\sqrt{n}}<a^Tp(x)<b_{k+2}-\frac{1}{4\sqrt{n}}\}$, $k\in[0:\prod_{s=1}^{l}k_s-1]$,
\begin{align}
&\mathbb{P}\left(y=y_k \mid \mathcal{X}^k\right) = 1 \text{ for some } y_k \in \mathcal{Y}\label{eqn:con11},\\
&\mathbb{P}\left(\cup_{k}\mathcal{X}^k\right)= 1.\label{eqn:con12}
\end{align}

The equation~\eqref{eqn:rel1} implies that $\mathbb{P}(\mathcal{X}^{\tilde{k}(\mathbf{i})}|\mathcal{X}_{\mathbf{i}})=1$.
Furthermore, we can show that $\mathbb{P}(\cdot\mid\mathcal{X}^{\tilde{k}(\mathbf{i})})=\mathbb{P}(\cdot\mid\mathcal{X}_{\mathbf{i}})$, since
\beq
\begin{split}
\mathbb{P}\left(\mathcal{X}_{\mathbf{i}}|\mathcal{X}^{\tilde{k}(\mathbf{i})}\right) 
& =1{-}\sum_{\mathbf{j}\neq\mathbf{i}}\mathbb{P}(\mathcal{X}_{\mathbf{j}} | \mathcal{X}^{\tilde{k}(\mathbf{i})})\\
& =1{-}\sum_{\mathbf{j}\neq\mathbf{i}}\frac{\mathbb{P}(\mathcal{X}_{\mathbf{j}})}{\mathbb{P}(\mathcal{X}^{\tilde{k}(\mathbf{i})})}\mathbb{P}(\mathcal{X}^{\tilde{k}(\mathbf{i})} \mid \mathcal{X}_{\mathbf{j}})=1
\end{split}
\eeq
where the first equality is from the fact $\mathcal{X}_{\mathbf{j}}$ does not overlap and partitions $\mathcal{X}$, and the last one is from the fact $\tilde{k}(\mathbf{i})\neq\tilde{k}(\mathbf{j})$ since $\tilde{k}$ is injective. 

There exists $\mathbf{i}_k\in\prod_{s=1}^{n}[1:k_s]$ such that $k=\tilde{k}(\mathbf{i}_k)$ since $\tilde{k}$ is surjective. Then, from  the condition (i) of Def.~\ref{def:sep2}, 
 \(\mathbb{P}_{(x,y)\sim D}\left(y=y_{\mathbf{i}_k} \mid \mathcal{X}_{\mathbf{i}_k}\right) = 1\) for some \(y_{\mathbf{i}_k} \in \mathcal{Y}\).
By using this and $\mathbb{P}(\cdot\mid\mathcal{X}^{\tilde{k}(\mathbf{i})})=\mathbb{P}(\cdot\mid\mathcal{X}_{\mathbf{i}})$, the first condition~\eqref{eqn:con11} can be proven where $y_k=y_{\mathbf{i}_k}$.

The second condition~\eqref{eqn:con12} can also be shown by
\beq
\begin{split}
\mathbb{P}\left(\bigcup_{k}\mathcal{X}^k\right) & =\sum_{\mathbf{j}}\mathbb{P}\left(\bigcup_{k}\mathcal{X}^k \Big| \mathcal{X}_{\mathbf{j}}\right)\mathbb{P}(\mathcal{X}_{\mathbf{j}})\\
& \geq\sum_{\mathbf{j}}\mathbb{P}\left(\mathcal{X}^{\tilde{k}(\mathbf{j})} \big| \mathcal{X}_{\mathbf{j}}\right)\mathbb{P}(\mathcal{X}_{\mathbf{j}})=1
\end{split}
\eeq
since $\mathbb{P}\left(\mathcal{X}^{\tilde{k}(\mathbf{j})} \big| \mathcal{X}_{\mathbf{j}}\right)=1$ and $\sum_{\mathbf{j}}\mathbb{P}(\mathcal{X}_{\mathbf{j}})=1$ by the condition (ii) of Def.~\ref{def:sep2}.

Thus, \(D'\) satisfies all conditions to be  \((\prod_{s=1}^{n}k_s)\)-separable with \(\frac{1}{4\sqrt{n}}\)-margin by a projection vector $a=\frac{1}{\sqrt{n}}[1,1,\dots,1]^T\in\mathbb{R}^n$. \IEEEQED




\bibliographystyle{IEEEtran}

\end{document}